\documentclass[10pt,twocolumn,letterpaper]{article}

\usepackage[pagenumbers]{wacv} % To force page numbers, e.g. for an arXiv version

\usepackage{graphicx}
\usepackage{amsmath}
\usepackage{multirow}
\usepackage{amssymb}
\usepackage{dsfont}
\usepackage{booktabs}

\usepackage[pagebackref,breaklinks,colorlinks]{hyperref}

\usepackage[capitalize]{cleveref}
\crefname{section}{Sec.}{Secs.}
\Crefname{section}{Section}{Sections}
\Crefname{table}{Table}{Tables}
\crefname{table}{Tab.}{Tabs.}

\def\wacvPaperID{2730} % *** Enter the WACV Paper ID here
\def\confName{WACV}
\def\confYear{2025}

\begin{document}

\title{DT-LSD: Deformable Transformer-based Line Segment Detection}

\author{Sebastian Janampa\\
	The University of New Mexico\\
	{\tt\small sebasjr1966@unm.edu}
	\and 
	Marios Pattichis\\
	The University of New Mexico\\
	{\tt\small pattichi@unm.edu}
%\and
%Marios Pattichis\\
}
\maketitle

%%%%%%%%% ABSTRACT
\begin{abstract}
  Line segment detection is a fundamental low-level task in computer vision, and improvements in this task can impact more advanced methods that depend on it. Most new methods developed for line segment detection are based on Convolutional Neural Networks (CNNs). Our paper seeks to address challenges that prevent the wider adoption of transformer-based methods for line segment detection. More specifically, we introduce a new model called Deformable Transformer-based Line Segment Detection (DT-LSD) that supports cross-scale interactions and can be trained quickly. This work proposes a novel Deformable Transformer-based Line Segment Detector (DT-LSD) that addresses LETR's drawbacks. For faster training, we introduce Line Contrastive DeNoising (LCDN), a technique that stabilizes the one-to-one matching process and speeds up training by 34$\times$. We show that DT-LSD is faster and more accurate than its predecessor transformer-based model (LETR) and outperforms all CNN-based models in terms of accuracy. In the Wireframe dataset, DT-LSD achieves 71.7 for $sAP^{10}$ and 73.9 for $sAP^{15}$; while 33.2 for $sAP^{10}$ and 35.1 for $sAP^{15}$ in the YorkUrban dataset. Code available at: \url{https://github.com/SebastianJanampa/DT-LSD}.
\end{abstract}

\section{Introduction}
\label{sec:intro}

\begin{figure*}[t]
	\centering
	\resizebox{\linewidth}{!}{
		\begin{tabular}{ccc}
			\phantom{a}\includegraphics[width=0.9\linewidth]{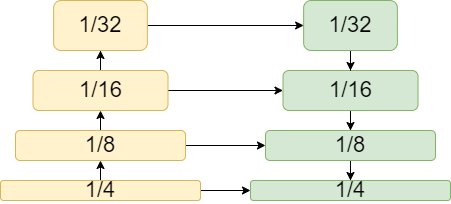}\phantom{a}& \phantom{a}\includegraphics[width=0.9\linewidth]{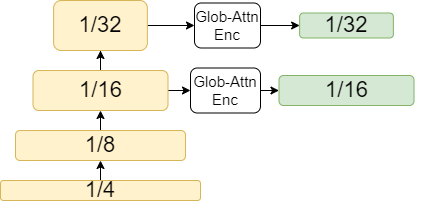}\phantom{a}& \phantom{a}\includegraphics[width=0.9\linewidth]{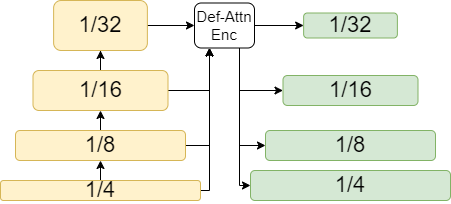}\phantom{a}\\
	\end{tabular}}
	\resizebox{\linewidth}{!}{
		\begin{tabular}{ccc}
			\phantom{hello}a) Feature Pyramid Network\phantom{hello} & \phantom{hello}b) Global Attention Encoder\phantom{hello} & \phantom{hello}c) Deformable Attention Encoder\phantom{hello}
	\end{tabular}}
	\caption{Feature map enhancing. All line segment detectors use a hierarchical backbone, but they differ from each other in their enhancing method. (a) CNN-based models use a feature pyramid network to combine two contiguous feature maps, allowing the propagation of global information to low-dimensional feature maps. However, no intra-scale interaction is applied to any feature map. 
		(b) LETR \cite{LETR} uses a global attention encoder for each processed feature map, promoting the intra-scale interaction but not the cross-scale interaction since no information is passed between the two processed feature maps. 	
		(c) DT-LSD allows intra- and cross-scale (more than two feature maps) interactions by applying a deformable-attention encoder.}
	\label{fig: overview}
\end{figure*}

Line segment detection is a low-level vision task used for higher-level tasks such as 3D reconstruction, camera calibration, vanishing point estimation, and scene understanding. Despite its importance, this problem remains open. Additionally, unlike other computer vision tasks (\eg object detection, 3d-estimation, camera calibration), transformer-based models are not popular to tackle this challenge,  LinE segment TRansformers (LETR) \cite{LETR} is the only transformer model for line segment detection in the literature. Most recent methods \cite{hawp, lcnn, mlnet, sacwp, HAWPv2} uses Convolutional Neural Networks (CNN) despite the fact that CNNs require a post-processing step to get the final predictions.

All models have a backbone that produces a set of hierarchical feature maps for further processing as shown in \cref{fig: overview}. CNN-based models use a feature pyramid network (FPN) as an enhancing method following the HourglassNet method \cite{hourglass} (see \cref{fig: overview}a). This method demonstrates the importance of cross-scale interaction since the new feature map is computed from contiguous feature maps, allowing the propagation of the global information from the highest-level feature map to the lower-level ones. On the other hand, LETR produces an enhanced feature map using a single feature map, as depicted in \cref{fig: overview}b. LETR demonstrates the ability of the global attention mechanism \cite{attention} to capture long-term relationships between the pixels of the same feature map (intra-scale processing), providing rich features. 

This paper develops  a  new transformer-based models for line segment detection. First, we improve LETR's feature map-enhancing method. We choose the deformable attention mechanism \cite{deformable-DETR} for its ability to combine both intra- and cross-scale processing. We illustrate our idea in \cref{fig: overview}c, where a deformable-attention encoder is used for feature map enhancement. The encoder receives a set of hierarchical feature maps\footnote{feature maps are pre-processed by a $1\times1\text{\phantom{a}}\mathrm{conv}$ to assure all the inputs have the same amount of channels.} where for a pixel a fixed number of sampling offsets are generated for each given feature map. Second, we reduce the number of epochs required for training. 
%As shown in \cite{deformable-DETR,dab-DETR,dn-DETR}, applying multiple feature maps accelerates convergence; however, end-to-end transformer-based models suffer from an unstable Hungarian matcher. 
Inspired from \cite{dn-DETR, dino}, we propose Line Contrastive DeNoising (LCDN) as training technique  to to accelerate the convergence of the training process. We show the efficiency of LCDN in \Cref{tab:dn_components} where we improve the metrics while keeping the same amount of epochs.  

Our contributions are summarized as follows:
\begin{enumerate}
	\item  We propose a novel end-to-end transformer-based framework showing that outperforms CNN-based line segment detectors. This is achieved by using the deformable attention mechanism.
	\item We introduce a highly-efficient training technique, Line Contrastive DeNoising, to reduce the number of epochs. This technique allows DT-LSD to achieve convergence in a similar number of epochs to CNN-based models.
	\item On two datasets (Wireframe \cite{wireframe_data} and YorkUrban \cite{york_data}), our end-to-end  transformer-based model present a performance improvement over state-of-the-art methods on both structural and  heat map metrics. 
	\item Our work opens up opportunities for line segment detectors to remove hand-crafted post-processing by utilizing end-to-end transformer-based models.
\end{enumerate}

In what remains of this paper, we describe previous state-of-the-art methods in line segment detection, as well as the two attention mechanisms in \cref{sec: background}. Next, we describe the methodology for DT-LSD in \cref{sec: methodology}. Then, we provide information about model parameters and training settings, comparison against previous state-of-the-art models, and the ablation studies of DT-LSD in \cref{sec: results}. Finally, we summarize our findings in \cref{sec: conclusion}.

\section{Background}
\label{sec: background}
\subsection{Line Segment Detection}
\subsubsection{Traditional Approaches}
The Hough Transform (HT) \cite{hough} remains an important method for line detection.
First, the Hough Transform applies Canny edge detection  \cite{canny} to obtain line segment candidates. Candidate lines are represented in polar form.
Here, we note that candidate lines are evaluated based on the overlapping number of pixels between the lines and the detected edges. Variations include the use of the Radon Transform and the Revoting Hough Transform. 

In regions dominated by a large density of edges, the HT can generate a large number of false positives. Grompone von Gioi \etal proposed a linear-time Line Segment Detector (LSD) \cite{LSD} to address this problem. LSD uses \textit{line-supported} regions and line segment validation. The approach also reduced time complexity
through the use of a pseudo-sorting algorithm based on gradient magnitudes.
A fundamental advantage of traditional methods	is that they do not require training for specific datasets.

\subsubsection{Deep Learning Based Approaches} 
%CNN models consist of a convolutional encoder-decoder as a backbone network to generate feature maps that are further processed to generate line segment estimates. Top-down strategies (\eg, \cite{hawp}) first use an attraction field map to detect possible line segment region candidates, then squeeze these regions into line segments to make a final prediction. On the other hand, bottom-up strategies \cite{afm,lcnn} detect line junctions, which are used to predict candidate line segments. Then, a classifier validates the candidates and produces the final set of predicted line segments. However, all of these models require of post-processing steps such as Non-Maximum Suppression (NMS) which is requires of multiple trial and errors to find the best value.

Learning-based line segment detectors have shown significant improvements compared to traditional approaches. The methods include different approaches that focus on line junctions, attraction field maps (AFM), transformers, and combining traditional approaches with deep learning techniques. 

The Holistically-Attracted Wireframe Parser (HAWP) \cite{hawp} proposed a 4-dimensional attraction field map, and later HAWPv2 \cite{HAWPv2}, a hybrid model of HAWP and self-supervision, was introduced. MLNet \cite{mlnet} and SACWP \cite{sacwp} incorporated cross-scale feature interaction on HAWP model. In \cite{afm,lcnn}, the authors developed a method for detecting line junctions which were used to provide candidate line segments. Then, a classifier validated the candidates and produced the final set of predicted line segments. LSDNet [18] used a CNN model to generate an angle field and line mask that were used to detect line segments using the LSD method. HT-HAWP and HT-LCNN \cite{lin2020ht} added global geometric line priors through the Hough Transform  in deep learning networks to address the lack of labeled data. However, the above methods requires of post-processing steps to produce the final output. In contrast, Line segment transformers (LETR) \cite{LETR} remove post-processing steps by using an end-to-end transformer-based model that relies on a coarse-to-fine strategy with two encoder-decoder transformer blocks. 

\subsection{Transformers}
\label{sec:transformers}
\subsubsection{High-complexity of Global Attention Models}
One crucial factor of LETR's slow convergence is the global attention mechanism, which only does intra-scale feature processing. Plus, the global attention leads to  very high computational complexity. 

To understand the complexity requirements of LETR, we revisit the global attention mechanism. We define global attention using \cite{attention}:
\begin{equation}
	\mathrm{GlobAtt}(Q, K, V) = \mathrm{softmax}\big(\frac{QK^T}{\sqrt{d}}\big)V
	\label{eq:glob_attn}
\end{equation}
where $Q, K, V, \text{ and }d$ represent the queries, keys, values and the hidden dimensions, respectively. For object and line segment detection, we define $K=V \in \mathbb{R}^{HW\times d}$ as the flattened form of the feature map $f\in \mathbb{R}^{H\times W\times d}$ where $H$ and $W$ are the height and width, respectively. In the encoder, we have $Q=K=V$ resulting in a complexity time of $O(H^2W^2d)$. Similarly, for the decoder, we have $Q\in \mathbb{R}^{N\times d}$ where $N$ is the number of queries producing a complexity time of $O(HWC^2 + NHWC)$.

\subsubsection{Deformable Attention Module}
Based on our previous discussion, it is clear that the  bottleneck of transformer-based models is the encoder, whose complexity time quadratically increases with respect to the spatial size of the feature map. 
%In addition, the models required lots of memory, making high-dimensional feature maps infeasible for real-time applications. 
To address this issue, Zhu \etal \cite{deformable-DETR} proposed the deformable attention mechanism, inspired by deformable convolution \cite{deform_conv}. Unlike global attention, the deformable attention module only attends to a fixed number $k$ of keys for each query (see Fig. 2 from \cite{deformable-DETR} for a visual representation).

Given an input feature map ${f} \in \mathbb{R}^{H\times W\times d}$, let $q$ be the index of a query element with content feature $z_q$ and a 2d-reference point $p_q$, the deformable attention for one attention head\footnote{The multi-head deformable attention equation is in page 5 section 4.1 in \cite{deformable-DETR}} is mathematically defined as
\begin{equation}
	\mathrm{DeformAttn}(z_q, p_q, x) =  \sum_{i=1}^{k} A_{qi} \cdot f(p_q + \Delta p_{qi})
	\label{eq:deform_attn}
\end{equation}
where $i$ indexes the sampling keys and $k$ is the total sampling keys number ($k\ll HW$). The $i^{th}$ sampling key is computed as $p_q + \Delta p_{qi}$ where $\Delta p_{qi}$ is the sampling offset. $A_{qi}$ is the $i^{th}$ row of the attention weight $A_q \in \mathbb{R}^{k\times d}$. The weights of $A_{qi}$ satisfy $\sum_{i=1}^{k}A_{qi} = 1$. 

Comparing \cref{eq:glob_attn} to \cref{eq:deform_attn}, $\mathrm{softmax}\big(QK^T/\sqrt{d}\big)$ is replaced by $A_q$, and $V$ by $x(p_q + \Delta p_{qi})$. So, the complexity time for the deformable encoder is $O(HWd^2)$, which has a linear complexity. For the deformable decoder, the complexity time is $O(kNd^2)$ where $N$ is the total number of queries, and the spatial dimensions of $x$ are irrelevant. 

Apart from reducing the memory and time complexities, deformable attention has a variation called multi-scale deformable attention ($\mathrm{MSDeformAttn}(z_q, \hat{p}_q, \{f_l\}_{l=1}^L)$) that allows cross-scale feature interaction and is defined as

\begin{equation}
	\sum_{l=1}^{L}\sum_{i=1}^{k}A_{lqi}\cdot f_l(\phi_l(\hat{p}_q) + \Delta p_{lqi})
	\label{eq:multi_deform_attn}
\end{equation}

where $\{f_l\}_{l=1}^L$ is a set of $L$ multi-scale feature maps, and $f_l\in \mathbb{R}^{H_l\times W_l \times d}$. The normalized 2d coordinates $\hat{p}_q$ has its  values lying in the range of $[0, 1]$, and is re-scaled to dimensions of the $l^{\text{th}}$-level feature map by the function $\phi_l$. Like \cref{eq:deform_attn}, the attention weight $A_{lqi}$ satisfies $\sum_{l=1}^{L}\sum_{i=1}^{k}A_{lqi} = 1$.

\section{Methodology}
\label{sec: methodology}
%\subsection{Why FT-LSD?}
%Vanilla DETR does not work very well in detecting small objects \cite{review1, review2} because it uses the feature map from ResNet's last stage, which downsamples the input image by a factor of 32 or 16 (for DETR-DC5). While reducing the image's dimensions speeds up procession and lowers memory consumption, it results in a loss of spatial information, especially for small objects. 
%
%When it comes to thin lines, reducing the dimensions of an image will lead to a loss of line geometry information. LETR \etal \cite{LETR} addresses this issue by using a coarse-to-fine-technique. LETR has two transformer blocks for coarse and fine detection that receive feature maps from ResNet \cite{resnet} stages 3 and 4, respectively. However, LETR does not do cross-scale feature procession. Instead, it passes coarse line queries to the second transformer block to produce more refined line segments (see \cref{fig: overview}b).
\begin{figure*}[!t]
	\centering
	\includegraphics[width=0.9\textwidth]{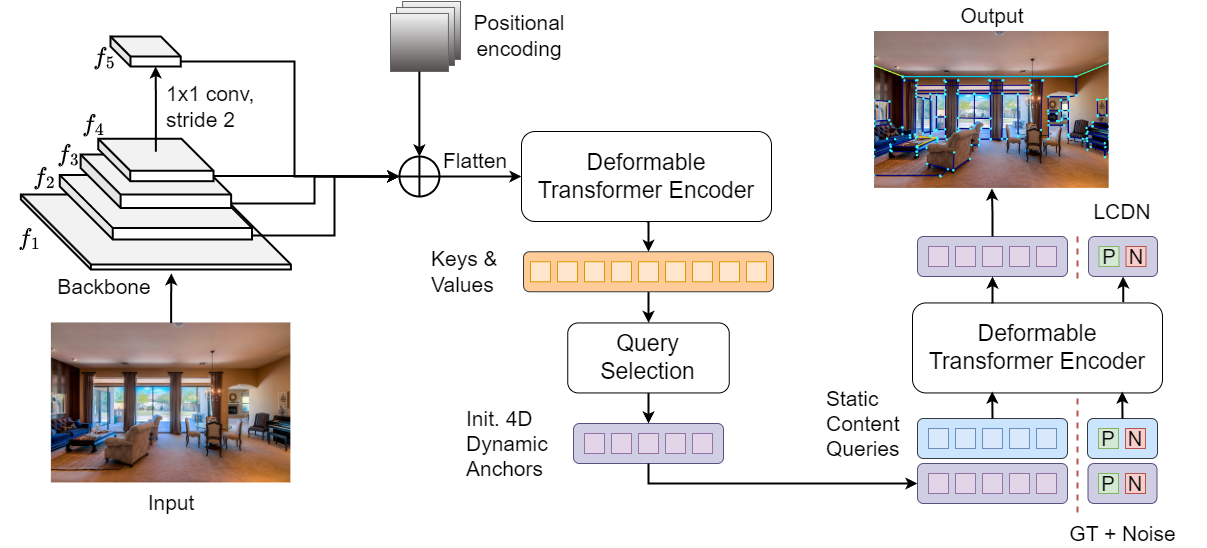}
	\caption{Framework of the proposed DT-LSD model.
		DT-LSD uses a deformable encoder and deformable decoder layers.
		Furthermore, it uses a set of mixed queries as a training strategy which does not influence the inference time.}
	\label{fig: model}
\end{figure*}

\subsection{Overview}

We present the architecture of DT-LSD, an end-to-end deformable transformer for line segment detection, in \cref{fig: model}. First, we pass an RGB image to a backbone to produce a set of hierarchical feature maps. Second, a deformable encoder enhances the backbone's feature maps. Third, we apply query selection to choose the top-K queries \footnote{In the encoder, feature maps pixels are treated as queries} as the initial 4D dynamic line endpoints. Fourth, we feed the initial dynamic line endpoints and the static (learnable) content queries to the deformable decoder to promote the interaction between queries and the enhanced feature maps. Fifth, two independent multi-layer perceptron networks process the decoder's output queries to estimate the line segment endpoints and classify whether a query contains a line. For the training process, we added an extra branch to perform line contrastive denoising, which did not affect the inference time. In this section, we do not describe the decoder and the one-to-one matching since they are already described in \cite{deformable-DETR} and \cite{LETR}.

%In what follows of this section, we will describe only the line contrastive denoising technique since the other components of our model are already described in \cite{deformable-DETR, dab-DETR, dino, dn-DETR}. For the 4D dynamic line endpoints, we change the representation of 4D dynamic anchors given by Dynamic-Anchor-Boxes (DAB) DETR \cite{dab-DETR} and use the $xyxy$ notation instead of the $xywh$ notation. For the Hungarian Matching, we use the same as in LETR \cite{LETR}.

\subsection{Deformable Transformer Encoder}
\begin{figure}
	\includegraphics[width=0.95\columnwidth]{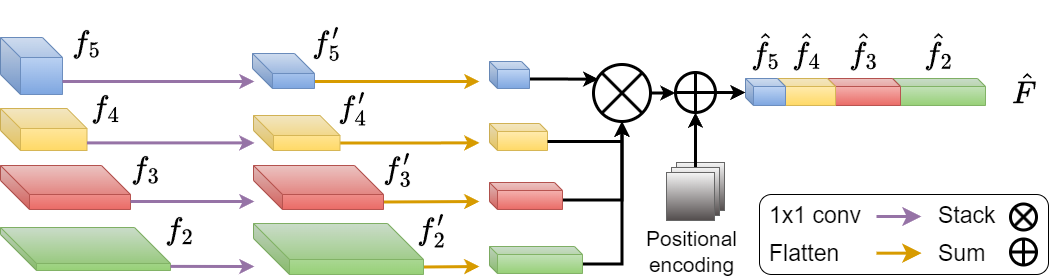}
	\caption{Feature maps pre-processing for the encoder.}
	\label{fig:feat_pre_proc}
\end{figure}

The encoder is a fundamental part of our network since it enhances the backbone's feature maps. However, these feature maps do not have any dimensions in common. For this reason, it is important to pre-process them before passing them to the encoder. As shown in \cref{fig:feat_pre_proc}, given an RGB image of dimensions $(H, W, 3)$, the backbone produces a set of hierarchical feature maps $\{f_l\}_{l=1}^5$ where $f_l\in \mathbb{R}^{H_l\times W_l \times d_i}$ and $H_l = H/2^{l+1}$ and $W_l = W/2^{l+1}$. Since $\{f_l\}_{l=1}^5$ do not have any dimension in common and we do not want to lose spatial resolution, we apply a $1\times1$ convolution to each feature so that  the whole set has the same number of channels $\{f'_l\}_{l=2}^5$ where $f'_l\in \mathbb{R}^{H_l\times W_l \times 256}$. Next, we  flatten their spatial dimension, followed by stacking them together and adding the position encoding ($PE$), creating the vector 
\begin{equation}
	\hat{F} = \mathrm{stack}(\hat{f}_2, \hat{f}_3, \hat{f}_4, \hat{f}_5)
\end{equation}
where $\hat{f}_i = \mathrm{flatten}(f'_i + PE(f'_i))$ and $\hat{F}\in\mathbb{R}^{L \times 256}$ \footnote{The $\mathrm{stack}$ and $\mathrm{flatten}$ functions are associative functions.}, $L = \sum_{i=1}^5 H_i\cdot W_i$.

We pass $\hat{F}$ to the encoder where each of its stacked pixels is treated as a query $q \in \mathbb{R}^{1\times256}$. For each query, we produce a fixed number $k=4$ of offsets per feature map followed by applying \cref{eq:multi_deform_attn}. In general, we compute a total of 16 offsets per attention head, where 4 of them are for intra-scale interaction, and the other 12 are for cross-scale interaction. Besides combining both types of interactions, we address the global attention's time complexity and the convolutional layers' kernel space restriction.

\subsection{Line Contrastive Denoising Technique}
\label{sec:lcdn}
 A problem for end-to-end transformer-based methods is the  one-to-one matching technique, which removes the need for non-maximum suppression (NMS) but is unstable in matching queries with ground truth. The main difference between one-to-one matching and NMS is that the first one uses scores to do the matching. In contrast, the second one eliminates candidates depending on how much the bounding boxes of the candidates belonging to the same class overlap. 

\begin{figure}[!t]\centering	
	\resizebox{0.98\linewidth}{!}{
	\begin{tabular}{ccc}
		\includegraphics[width=0.95\columnwidth]{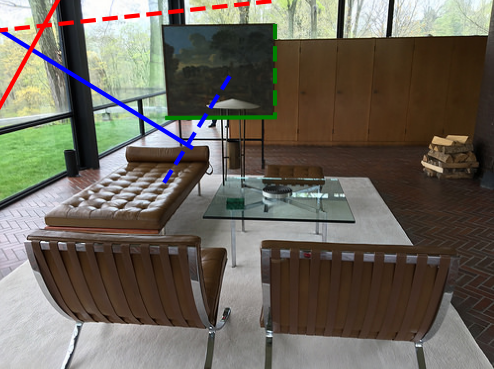} & \phantom{aaa}&
		\includegraphics[width=0.95\columnwidth]{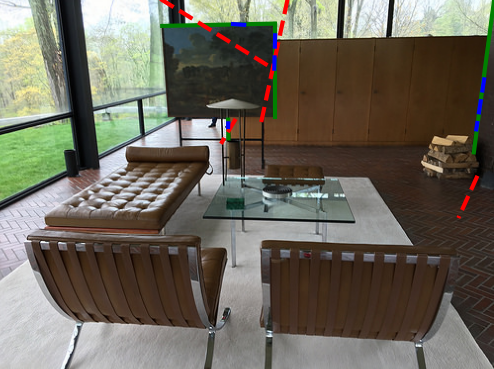}\\
	\end{tabular}
}
\resizebox{0.98\linewidth}{!}{
\begin{tabular}{ccc}
	\phantom{hello}a) CDN (DINO)\phantom{hello}& \phantom{aaa} & \phantom{hello}b) LCDN (ours)\phantom{hello}
\end{tabular}
}
	\newcommand{\crule}[3][red]{\textcolor{#1}{\rule{#2}{#3} \rule{#2}{#3} \rule{#2}{#3}}}
		\resizebox{0.98\linewidth}{!}{
		\begin{tabular}{ccc}
			\crule[green]{0.01\linewidth}{0.01\linewidth} Ground Truth & 
			\crule[blue]{0.01\linewidth}{0.01\linewidth} Positive query &
			\crule[red]{0.01\linewidth}{0.01\linewidth} Negative query 
		\end{tabular}
	}
	\caption{Comparison between contrastive denoising techniques applied to line segments. We present two different line segments and their positive and negative queries. We use solid and dashed to different between line segment samples.}
	\label{fig: lcdn}
\end{figure}

In this section, we present Line Contrastive Denoising (LCDN), a training technique for stabilizing the matching process inspired from \cite{dino}. While LCDN is used to speed up training, it is not part of the final inference model. LCDN facilitates the matching by teaching the Hungarian Matcher to accept queries whose predicted line's endpoints lie on or are close to a ground-truth line and to reject queries whose predicted line's endpoints are far away from the ground-truth line. To achieve this, we create positive and negative queries by performing line length scaling and line rotation. The length scaling consists of varying the length of the line segment, with original length ${l}$, such that positive queries have a length in a range of $[0, l]$ and the negative queries $(l, 2l)$. For line rotation, we rotate the line in a range of $(-\tau, \tau)$  for positive, where $\tau$ is the fixed angle. For negative queries, the rotation is in the $(-2\tau, -\tau] \cup [\tau, 2\tau)$ range. We present an example of our LCDN technique in \cref{fig: lcdn}b and a comparison against the Contrastive DeNoising (CDN)  \cite{dino} in \Cref{tab:dn_components}.

Since LCDN generates extra groups of denoising queries from ground-truth lines, this can harm the training process if the prediction queries interact with the denoising queries. The denoising queries contain ground-truth information, so if a matching query sees this information, it will start training with information that should not be known. In the other case, we want the denoising queries to see the information stored by the matching queries. We manually implement this by using an attention mask. Note that only queries from the same denoising group are allowed to interact with each other, but all denoising groups interact with the matching group.

%\begin{figure}
%	\centering
%	\includegraphics[width=0.6\columnwidth]{figures/algorithm/attention_mask}
%	\label{fig:attn_mask}
%\end{figure}

\subsection{Loss Function}

We choose the focal loss \cite{focal_loss} for line classification 
because it can deal with class imbalance. 
The focal loss encourages training on uncertain samples by penalizing samples with predicted probability $\hat{p}$ that is away from 0 or 1 
as given by (per sample loss):
\begin{equation}
	\mathcal{L}_{\text{class}}^{(i)} = -(\alpha_1(1-\hat{p}^{{(i)}})^\gamma\log \hat{p}^{(i)}+\alpha_2 (\hat{p}^{{(i)}})^\gamma\log (1-\hat{p}^{{(i)}}))
\end{equation}
with $\alpha_1=1$, $\alpha_2=0.25$, and $\gamma=2$.

For each line candidate $\hat{\mathbf{l}}$, we use the $L_1$ loss  to compute the distance 
from the ground truth points.
Let $i$ denote the $i^{th}$ line candidate.
If the classifier accepts the line candidate, we get $c_i=1$ for the classification output.
Else, $c_i=0$.
Let $\mathbf{l}^{(i)}_j$ denote the $j^{th}$ endpoint component of the $i$-th ground-trtuh line.
We have four endpoint components because we use two coordinate
points representing the line segment.
The $\mathcal{L}_{\text{line}}$ loss function is then given by:       
\begin{equation}
	\mathcal{L}_{\text{line}}^{(i)} = \mathds{1}_{\{c_i \neq \emptyset\}}\sum_{j=1}^{4} |\mathbf{l}_j^{{(i)}} - \hat{\mathbf{l}}^{(i)}_j|.
\end{equation}
where $\mathds{1}_{\{c_i \neq \emptyset\}}$ is the indicator function based on the classifier output $c_i$.

The final loss $\mathcal{L}$ is a linear combination 
of the two loss functions:
\begin{equation}
	\mathcal{L} = \sum_{i=1}^{N}
	\lambda_{\text{cls}} \, \mathcal{L}_{\text{class}} ^{(i)}+ 
	\lambda_{\text{line}}\, \mathcal{L}_{\text{line}} ^{(i)}
\end{equation}
where $\lambda_{\text{class}}=2$, 
$\lambda_{\text{line}}=5$, and $N$ is the total number of instances.

\begin{table}
	\centering
	\resizebox{0.65\columnwidth}{!}{
		\begin{tabular}{l|l}
			\toprule
			Parameter & Value\\
			\midrule
			% Architecture
			number of feature maps & 4\\
			number of encoder layers & 6\\
			encoder sampling points & 4\\
			number of decoder layers & 6\\
			decoder sampling points & 4\\
			hidden dim & 256\\
			feedforward dim & 1024\\
			number of heads & 8\\
			number of classes & 2\\
			number of queries & 900\\
			
			%LCDN
			denoising number & 300\\
			label noise ratio & 0.5\\
			line scaling & 1.0\\
			line rotation & 7$^{\circ}$\\
			
			% Losses
			line loss weight & 5\\
			class loss weight & 2\\
			
			% hypeparams
			optimizer & AdamW\\
			initial learning rate & 1e-4\\
			initial learning rate of backbone & 1e-5\\
			weight decay & 1e-4\\
			batch size & 2\\
			total number of epochs & 24\\
			learning rate drop & 21\\
			\bottomrule
	\end{tabular}}
	\caption{DT-LSD architecture parameters and training setup.}
	\label{tab: hyperparameters}
\end{table}

\section{Results}
\label{sec: results}
\subsection{Datasets}
\label{sec:experiments}
\textbf{ShangaiTech Wireframe Dataset.}
We used the ShangaiTech Wireframe dataset for comparisons \cite{wireframe_data}. The dataset consists of 5462 (indoor and outdoor) images of hand-made environments. The ground-truth line segments were manually labeled. 
The goal of the dataset was to provide line segments with meaningful geometric information about the scene. We split the dataset into 5000 images for training and 462 for testing.

\textbf{York Urban Dataset.} 
We also use the York Urban dataset \cite{york_data}.
The dataset consists of 122 (45 indoor and 57 outdoor) images of size $640 \times 480$ pixels. Denis \etal generated ground-truth line segments using an interactive MATLAB program with a sub-pixel precision. We only used this dataset for testing.

For the results shown in \cref{sec: sota_comparison}, we follow \cite{dab-DETR, detr, deformable-DETR, dino, dn-DETR} and apply data augmentations to the training set.
%, including random horizontal/vertical flip, random resize, random crop, and image color jittering
During training, we  reprise the input images such that the shortest side is at least 480 and at most 800 pixels while the longest at most 1333. At the evaluation stage, we resize the
image with the shortest side at least 640 pixels.

\begin{table*}[t]
	\centering
	\resizebox{0.98\linewidth}{!}{
		\begin{tabular}{lrcrccccccrccccccrc}
			\toprule
			\multirow{2}{*}{Method} & \multirow{2}{*}{\phantom{a}} &  {\multirow{2}{*}{Epochs}} & \multirow{2}{*}{\phantom{a}} & \multicolumn{6}{c}{\textit{Wireframe Dataset}} &\multirow{2}{*}{\phantom{a}} & \multicolumn{6}{c}{\textit{YorkUrban Dataset}} & \phantom{a} & \multirow{2}{*}{FPS}\\
			\cmidrule{5-10} \cmidrule{12-17} 
			& & & & {sAP$^{10}$}  & {sAP$^{15}$}  & {\phantom{A}sF$^{10}$}   & {\phantom{A$^5$}sF$^{15}$}   & \phantom{s$^1$}AP$^{\text{H}}$   & \phantom{s}F$^{\text{H}}$  && {sAP$^{10}$}  & {sAP$^{15}$}  & {\phantom{A}sF$^{10}$}   & {\phantom{A$^5$}sF$^{15}$}   & \phantom{s$^1$}AP$^{\text{H}}$   & \phantom{s}F$^{\text{H}}$ \\
			\midrule
			\multicolumn{17}{l}{Traditional methods}\\
			LSD \cite{LSD} & & / & & / & /& / & / &55.2 & 62.5 & & / &/ &/ &/ &  50.9 & 60.1 && \textbf{49.6} \\
			\midrule 
			\multicolumn{17}{l}{CNN-based methods} \\
			DWP \cite{wireframe_data} && 120 && 5.1 & 5.9 & / & / & 67.8 & 72.2 && 2.1 & 2.6 & / & / & 51.0 & 61.6 && 2.24 \\
			AFM \cite{afm} && 200 && 24.4 & 27.5 & /& / & 69.2 & 77.2 && 9.4 & 11.1 & /& / & 48.2 & 63.3 && 13.5\\
			L-CNN \cite{lcnn} & & 16 &  & 62.9 & 64.9 & 61.3 & 62.4 & 82.8 & 81.3 & & 26.4 & 27.5 & 36.9 & 37.8 & 59.6 & 65.3 & & 10.3\\
			HAWP \cite{hawp} & & 30 & & 66.5 & 68.2 & 64.9 & 65.9 & 86.1 & 83.1 & & 28.5 & 29.7 & 39.7 & 40.5 & 61.2 & 66.3 & & {30.3}\\
			F-Clip \cite{fclip}&& 300 && 66.8 & 68.7 & / & / &85.1 & 80.9  && 29.9 & 31.3 & / &/ & 62.3 & 64.5 && 28.3 \\
			ULSD \cite{ulsd}&& 30 && 68.8 & 70.4 & / & / & / & / &&  28.8 & 30.6 & /& / & / & / && 36.8\\
			HAWPv2 \cite{HAWPv2} && 30 && 68.6 & 70.2 & / & / & $\underline{86.7}$ & 81.5 && 29.1 & 30.4 & / & /& 61.6 & 64.4 && 14.0 \\ 
			SACWP \cite{sacwp} && 30 && $\underline{70.0}$ & $\underline{71.6}$ & /& / &/ &/ && 30.0 & 31.8 &/ &/ &/ & / && 34.8 \\
			MLNET \cite{mlnet} && 30 && ${69.1}$ & ${70.8}$ & / & / & $\underline{86.7}$ & 81.4&& $\underline{32.1}$ & $\underline{33.5}$ & /& / & $\underline{63.5}$  & 65.1 && 12.6\\
			\midrule 
			\multicolumn{17}{l}{Transformer-based methods} \\
			LETR \cite{LETR} & & 825 & & 65.2 & {67.7} & $\underline{65.8}$ & $\underline{67.1}$ & 86.3 & $\underline{83.3}$ & & 29.4 & 31.7 & $\underline{40.1}$ & $\underline{41.8}$ & 62.7 & $\underline{66.9}$ & & 5.8\\  
			DT-LSD  (ours) && 24 && \textbf{71.7} & \textbf{73.9} & \textbf{70.1} & \textbf{71.2} & \textbf{89.1} & \textbf{85.8} && \textbf{33.2} & \textbf{35.1} & \textbf{44.5} & \textbf{45.8} & \textbf{65.9} & \textbf{68.0} && 8.9\\
			\bottomrule
		\end{tabular}
	}
	\caption{Line segment detection results.
		Based on the models trained on the Wireframe dataset, we provide
		test results on the both YorkUrban and Wireframe dataset.
		The \textbf{best} results are given in boldface.
		Underlines are used for the $\underline{\text{second best}}$.
	}
	\label{tab: results}
\end{table*}	

\begin{figure*}
	\centering
	\begin{tabular}{cccc}
		\includegraphics[width=0.48\columnwidth]{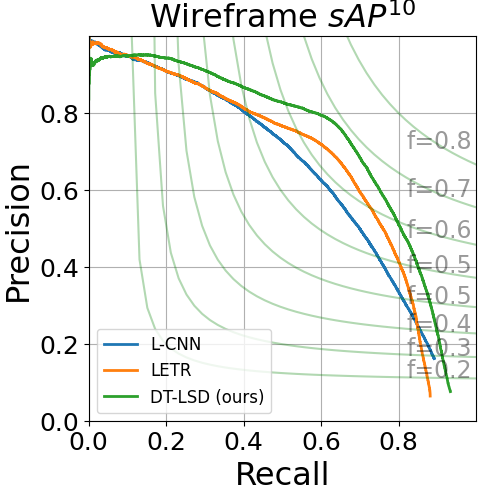}
		\includegraphics[width=0.48\columnwidth]{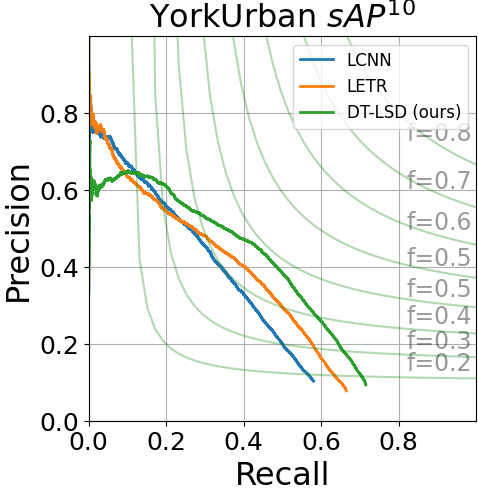}
		\includegraphics[width=0.48\columnwidth]{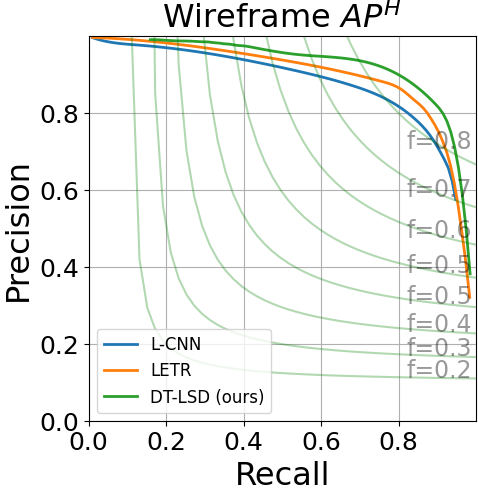}
		\includegraphics[width=0.48\columnwidth]{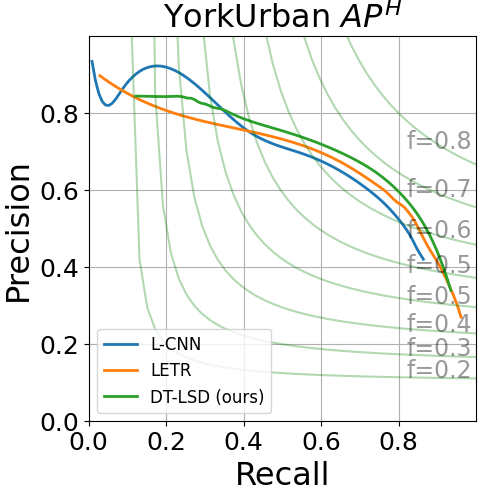}
	\end{tabular}
	\caption{Precision-Recall (PR) curves. PR curves comparisons between L-CNN\cite{lcnn}, LETR\cite{LETR} and DT-LSD(ours) using sAP$^{10}$ and AP$^\text{H}$ metrics for Wireframe and YorkUrban datasets.}
	\label{fig: pr_curve}
\end{figure*}

\subsection{Implementation}
\subsubsection{Network} 
We use the SwinL \cite{liu2021Swin} as a backbone for our deformable encoder-decoder transformer model. For the deformable transformer, we followed the recommendation of DINO \cite{dino}. We used 4 sampling offsets for the encoder and decoder, 900 queries to predict line segments, 6 stacked-encoder layers, and 6 stacked-decoder layers. We summarize the DT-LSD architecture parameters
and training parameters in \Cref{tab: hyperparameters}.
We train DT-LSD on a single Nvidia RTX A5500 GPU with a batch size of 2.

%??? I think it is best to add a simple paragraph.
%       Describe CPU version, Operating System, RAM, PyTorch or Tensorflow.
%       Also give model optimization parameters. ???

\subsection{Comparison to SOTA models}
\label{sec: sota_comparison}
We compare DT-LSD to many state-of-the-art models in \Cref{tab: results}. 
Our approach gives the most accurate result in all of our comparisons providing new state-of-the-art results in both datasets. In the Wireframe dataset, DT-LSD performs better than SCAWP (the second best result) by around 2 points in the sAP metric.
%  than LETR by 2.2 in the F$^\text{H}$. 
In the YorkUrban dataset, DT-LSD outperforms MLNET (the second best result) by around 1 percentage point in the sAP metric and 1.6 in the AP$^\text{H}$. Comparing to LETR, we show a significant gain in all metrics.
Furthermore, we note that DT-LSD is trained with just 24 epochs, 
significantly faster than the 825 epochs required for LETR.
At the same time, DT-LSD runs at nearly 9 frames per second.  

\begin{figure*}[!t]
	\centering
	\resizebox{0.98\linewidth}{!}{
		\begin{tabular}{ccccc}
			\includegraphics[width=0.25\linewidth]{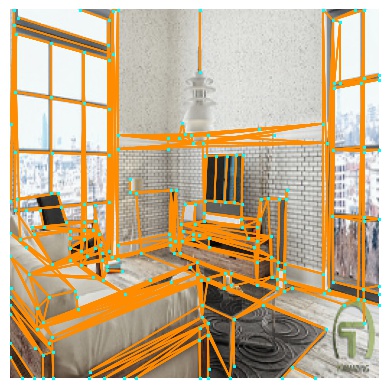} & 
			\includegraphics[width=0.25\linewidth]{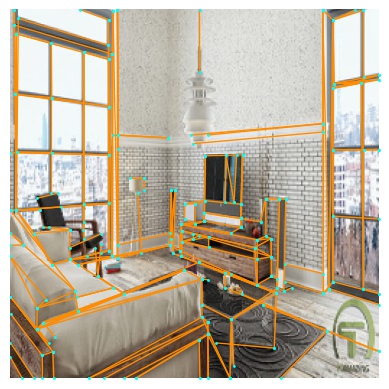} & 
			\includegraphics[width=0.25\linewidth]{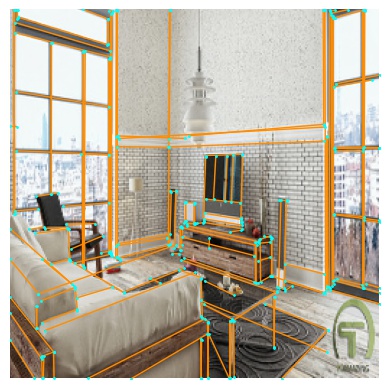} & 
			\includegraphics[width=0.25\linewidth]{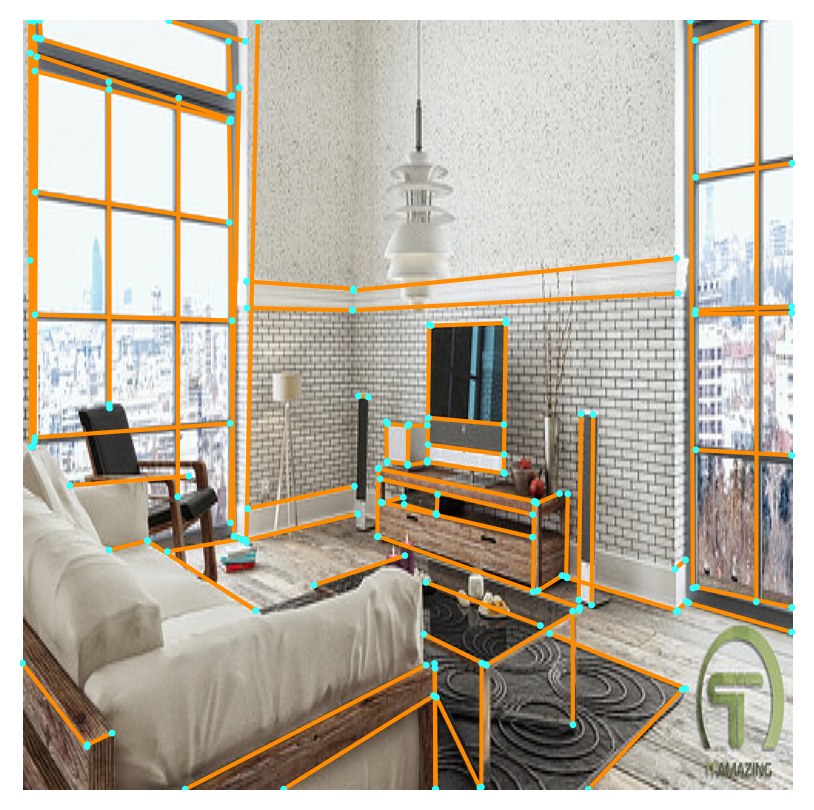} & 
			\includegraphics[width=0.25\linewidth]{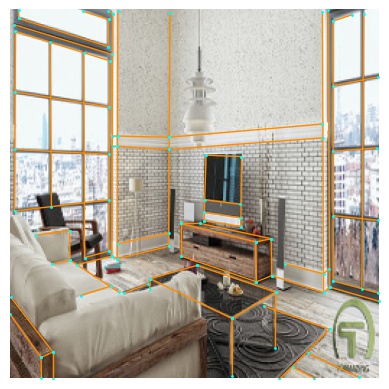}\\
			\includegraphics[width=0.25\linewidth]{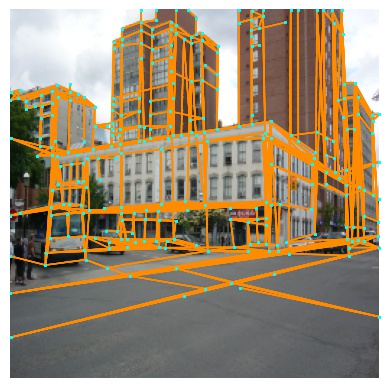} & 
			\includegraphics[width=0.25\linewidth]{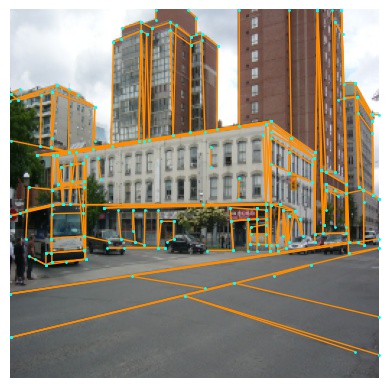} & 
			\includegraphics[width=0.25\linewidth]{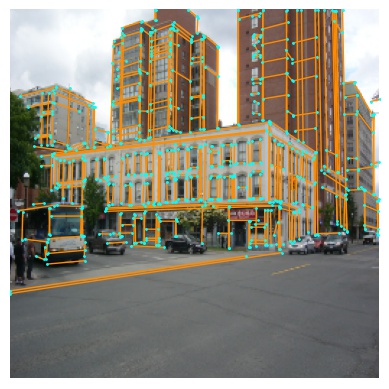} & 
			\includegraphics[width=0.25\linewidth]{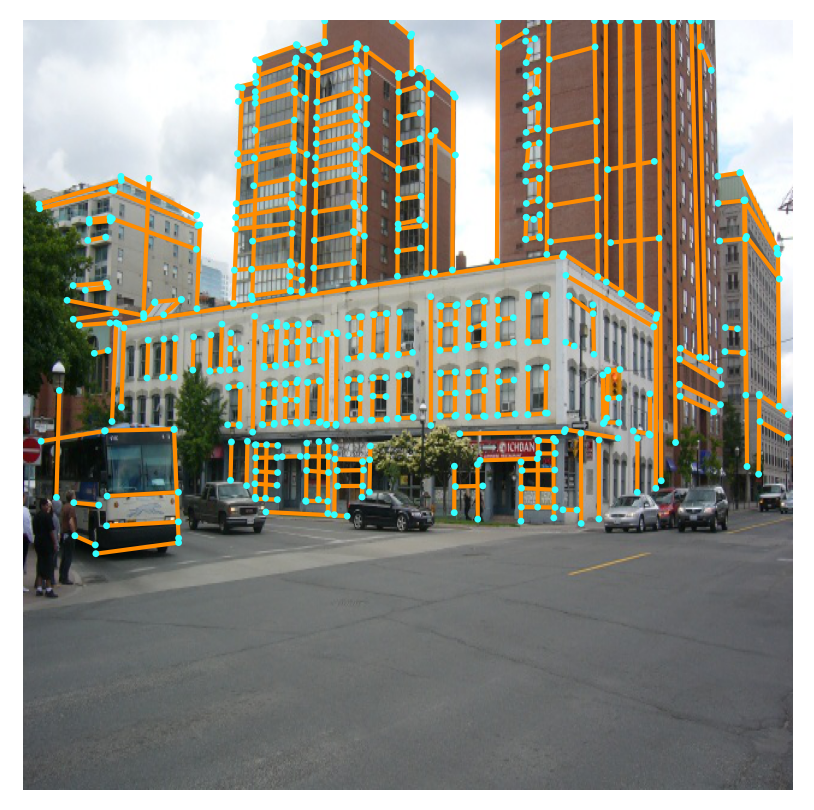} & 
			\includegraphics[width=0.25\linewidth]{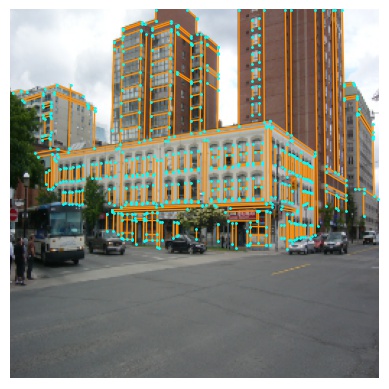}\\
			LCNN \cite{lcnn} & HAWP \cite{hawp} & LETR \cite{LETR} & DT-LSD (ours) & Ground-Truth
		\end{tabular}
	}
	\caption{Qualitative comparisons for line detection methods. 
		From left to right: the columns provide example results from 
		LCNN, HAWP, LETR, DT-LSD (ours) and the ground-truth. 
		The top and bottom rows provide examples from the Wireframe and YorkUrban test sets, respectively.
		The bottom two rows provide examples from the YorkUrban test set.}
	\label{fig: outputs}
\end{figure*}

We show the Precision-Recall (PR) curves for sAP$^{10}$ and AP$^\text{H}$ in \cref{fig: pr_curve}. DT-LSD has low performance for low recall values but surpasses LETR and L-CNN for recall values greater than 0.1. 
We also provide qualitative comparative results in \cref{fig: outputs}.  
The images clearly show that the transformer-based approaches (LETR and DT-LSD) perform significantly better than LCNN and HAWP. Upon closer inspection, we can see that DT-LSD avoids noisy line detections that appear in LETR
(\eg, inspect the center regions of third-row images). 

\subsection{Ablation Studies}
\subsubsection{Line Contrastive Denoising}

Our comparisons are based on the number of epochs and the sAP metric as summarized in   \Cref{tab:dn_components}. For a fair comparison, we train all the DINO\footnote{We use DINO-ResNet50-4scale from \cite{dino}.} variations using ResNet50 \cite{resnet} as the backbone. 

Compared to the 500 epochs required for the vanilla DETR, DINO converges at just 36 epochs.
Furthermore, plain DINO results in a maximum accuracy drop of 0.6, while our additions boost performance significantly (\eg, compare 66.3 and 68.8 versus 53.8 and 57.2).

The performance improvement against vanilla DETR is because DINO uses the MultiScale Deformable Attention mechanism described in \cref{eq:multi_deform_attn}, which promotes the cross- and intra-scale interaction. However, DINO has a lower performance than LETR. \cref{fig: lcdn} shows that the contrastive denoising (CDN) technique from DINO does not work for line segments because applying CDN to line segments results in scaling and translating both positive and negative queries. Therefore, these two operations lead to the model accepting non-line-segment candidates as potential candidates. We test our idea by removing the box scaling. We noticed an improvement of around 7 points for  sAP$^{10}$. We gain 3 extra points for sAP$^{10}$ by adding the line scaling technique, reaching a score of  $63.4$. By combining the line scaling and the line rotation (our line contrastive denoising technique), we obtain the maximum of $66.3$ in sAP$^{10}$. 

Based on our experiments, we conclude that our enhancing  method and training technique are effective since DINO with LCDN outperforms LETR using a lower-parameter backbone and fewer training epochs. We want to highlight that the input size of the image for all DINO variations was  $512\times512$ for both training and testing. On the other hand, vanilla DETR and LETR resize the images following the procedure described in \cref{sec:experiments} for training, and they resize the image with the shortest side of at least 1100 pixels for testing.

\begin{table}
	\centering
	\resizebox{\columnwidth}{!}{
		\begin{tabular}{l c  ccc}
			\toprule
			Model & Epochs & sAP$^{5}$ & sAP$^{10}$ & sAP$^{15}$\\
			\midrule
			Vanilla DETR & 500 & - & 53.8 & 57.2 \\
			LETR & 825 & - & 65.2 & 67.1 \\
			DINO &  36 &45.8 & 53.2 & 56.7\\
			\quad- box scaling &36& 51.7 & 60.0 & 63.5\\ 
			\quad+ line scaling & 36 & 56.5 & 63.4 & 66.2\\
			\quad+ line rotation & 36 & 60.7 & 66.3 & 68.7 \\
			%		+ angle noise:\\
			%		\qquad$3^\circ$  && 53.6 & 61.5 & 64.9\\
			%		\qquad$5^\circ$  && 53.8 & 61.8 & 65.1\\
			%		\qquad$10^\circ$  && 53.8 & 61.6 & 64.6\\
			%		\qquad$20^\circ$  && 54.1 & 61.5 & 64.6\\
			\bottomrule
		\end{tabular}
	}
	\caption{Ablation study of the different components of LCDN. All models uses ResNet50 as backbone except for LETR that uses ResNet101.}
	\label{tab:dn_components}
\end{table}	

\subsubsection{Feature maps}
An important element of our model is the feature maps generated by the backbone. Here, we evaluate the effectiveness of different feature maps and backbones.  We report the results in \Cref{tab: ablation_backbone}. All models are trained using 24 epochs. Adding the S1 feature map produces more precise line segments while slowing down our inference time (measured in the number of frames per second (FPS)). Using SwinL as the backbone gives the best results, but slows down the inference speed. For example, the configuration SwinL with 5 feature maps achieves a  score of $69.3$ in the sAP$^{10}$ at 10.5 FPS. %Adding the S1 feature map reduces the FPS from 13.6 fps down to 10.5 fps.

\begin{table}[t]
	\centering
	\resizebox{\columnwidth}{!}{
		\begin{tabular}{c c ccc c}
			\toprule
			Backbone & Feat. maps  & sAP$^{5}$ & sAP$^{10}$ & sAP$^{15}$ & FPS\\
			\midrule
			\multirow{2}{*}{ResNet50 \cite{resnet}}& S1- S5 & 60.2 & 65.5 & 68.1 & 13.0 \\
			& S2 - S5 & 59.5 & 65.1 & 67.5 & 18.5 \\
			\midrule
			\multirow{2}{*}{SwinL \cite{liu2021Swin}}& S1- S5 & 63.8 & 69.3 & 71.7 & 10.5 \\
			& S2 - S5 & 62.0 & 67.8 & 70.3 & 13.6\\
			\bottomrule
	\end{tabular}}
	\caption{Ablation study based on the number of feature maps and different backbones.}
	\label{tab: ablation_backbone}
\end{table}

\subsubsection{Image upsampling}
Most transformer-based algorithms use upsampling techniques to improve their performance. To evaluate the effects of  upsampling an image, we train and test our model at different resolutions for 24 epochs. As documented in \Cref{tab:ablation_upsampling}, upsampling benefits both CNN- and transformer-based models. First, we train DT-LSD following popular CNN-based methods by resizing the original image to 512$\times$512. 
For DT-LSD 512, we obtain the fastest
inference time at 13.6 FPS among all DT-LSD variations. % (??? faster than HAWP that don't have FPS??).

Motivated  by DETR-based models \cite{detr,deformable-DETR,dab-DETR,dn-DETR,LETR,dino,anchor-DETR}, we also apply scale augmentation consisting of resizing the input image so that the shortest side is a minimum of 480 pixels and a maximum of 800 pixels, while the longest side is a maximum of 1333 pixels. Here, we choose five different testing sizes, \textbf{1)} 480, the minimum size used for training, \textbf{2)}, 512, the size used for CNN-based line segment detectors, \textbf{3)} 640, the size use for YOLO detectors, \textbf{4)} 800, the maximum size used for training, and \textbf{5)} 1100, LETR's \cite{LETR} testing size. 

We note that this scaling technique improved our results over 512$\times$512.
For example, DT-LSD 480$^\dagger$ produces better results than DT-LSD 512. As the testing size increases, the sAP metric improves while reducing inference speed (as measured by FPS).  As a balance between speed and precision, we choose DT-LSD 640$^\dagger$ because it increases sAP$^{10}$ and sAP$^{15}$ by around 2 points, while its FPS is only by 2.9 fps less than DT-LSD 520$^\dagger$. 
We did not choose DT-LSD 800$^\dagger$ because the 0.4 improvement in sAP 
does not justify a drop of 2.5 fps in inference performance.

\begin{table}
	\centering
	\begin{tabular}{c c c cc c}
		\toprule
		Model & Train & Test & sAP$^{10}$ & sAP$^{15}$ & FPS\\
		& Size & Size & &  & \\
		\midrule
		HAWP & 512 & 512 & 65.7 & 67.4 & -\\
		
		HAWP & 832 & 832 &  67.7 & 69.1 & -\\
		
		HAWP & 832 & 1088 & 65.7 & 67.1 & -\\
		DT-LSD & 512 & 512 & 67.8 & 70.3 & 13.6\\
		DT-LSD & 800$^*$ & 480$^\dagger$ & 69.0 & 71.5 & 12.2\\
		DT-LSD & 800$^*$ & 520$^\dagger$ & 69.5 & 71.8 & 11.8\\
		DT-LSD & 800$^*$ & 640$^\dagger$ & 71.7 & 73.9 & 8.9\\
		DT-LSD & 800$^*$ & 800$^\dagger$ & 72.3 & 74.3 & 6.4\\
		DT-LSD & 800$^*$ & 1100$^\dagger$ & 72.2 & 74.2 & 4.7\\
		\bottomrule
	\end{tabular}
	\caption{Ablation study on the effects of image upsampling.
		We used square images. For 800$^*$, we process images with the smaller dimension
		between 480 and 800. 
		For the test sizes, we use $\mathrm{number}^\dagger$ to refer
		to the fact that the smaller dimension of the original image is given by
		$\mathrm{number}$.}
	\label{tab:ablation_upsampling}
\end{table}

\subsubsection{Training Epochs}
We report the results for training DT-LSD with 12, 24, and 36 epochs in \Cref{tab: ablation_epochs}. DT-LSD gets to competitive performance after training with just 12 epochs. At 12 epochs, DT-LSD achieves a sAP$^{10}$ of 68.4. There is very little difference between 24 and 36 epochs. At 36 epochs, we get a minor increase of 0.3, 0.2, and 0.1 in the  sAP$^{5}$, sAP$^{10}$, and sAP$^{15}$, respectively. 

\begin{table}
	\centering
	\begin{tabular}{c ccc}
		\toprule
		Number of Epochs & sAP$^{5}$ & sAP$^{10}$ & sAP$^{15}$\\
		\midrule
		12 & 65.2 & 68.4 & 69.7 \\
		24 & 66.6 & 71.7 & 73.9 \\
		36 & 66.9 & 71.9 & 74.0\\
		\bottomrule
	\end{tabular}
	\caption{Ablation study of the training schedule. DT-LSD trained using different numbers of epochs.}
	\label{tab: ablation_epochs}
\end{table} 

\subsubsection{Transfer Learning}
We report results based on pre-training on different datasets in \Cref{tab: ablation_pretrain}. For our experiments, we use 24 epochs. In our first example, the backbone was pre-trained with the  ImageNet-22k dataset \cite{imagenet22k}. In our second example, the DINO was pre-trained using the COCO object detection dataset \cite{coco}. From the results, it is clear that it is essential to pre-train the entire network and not just the backbone. 

\begin{table}
	\centering
	\begin{tabular}{c ccc}
		\toprule
		Pretrained Weights & sAP$^{5}$ & sAP$^{10}$ & sAP$^{15}$\\
		\midrule
		ImageNet-22k & 10.8 & 12.7 & 15.6\\
		COCO & 66.6 & 71.7 & 73.9 \\
		\bottomrule
	\end{tabular}
	\caption{Ablation study for pre-training using different datasets.}
	\label{tab: ablation_pretrain}
\end{table} 

\section{Conclusion}
\label{sec: conclusion}

We introduced DT-LSD, a transformer-based model for line segment detection.
DT-LSD uses cross-scale interactions to speed up convergence and improve results. 
Our approach uses pre-training on the COCO dataset to learn low-level features.
Our extensive experiment showed that end-to-end transformer-based model can surpass CNN-based methods. Additionally, we opened new opportunities for new line segment detection methods that do not require post-processing steps. 

In future work, we will consider the development of specialized backbones for transformer-based models. Additionally, an important observation from this work is that DT-LSD needs the COCO pre-trained weights to achieve state-of-the-art results. Therefore, we will also focus on the implementation of the network that is trained from scratch. 
%%%%%%%%% REFERENCES
\newpage
{\small
\bibliographystyle{ieee_fullname}
\bibliography{dtlsd}
}

\end{document}